\documentclass{article} 
\usepackage{iclr2022_conference,times}

\usepackage{amsmath,amsfonts,bm}

\def\eqref#1{equation~\ref{#1}}

\def\1{\bm{1}}

\DeclareMathAlphabet{\mathsfit}{\encodingdefault}{\sfdefault}{m}{sl}
\SetMathAlphabet{\mathsfit}{bold}{\encodingdefault}{\sfdefault}{bx}{n}

\usepackage{hyperref}
\usepackage{url}
\usepackage{graphicx}
\usepackage{placeins}

\title{\centering\begin{tabular}{c}
TEVR: Improving Speech Recognition\\
by Token Entropy Variance Reduction
\end{tabular}}

\author{%
\centerline{
Hajo Nils Krabbenhöft$^{\bigtriangleup}$, Erhardt Barth$^{\Box}$
} \vspace{1mm}\\
\centerline{\normalfont 
$^{\bigtriangleup}$~\begin{tabular}{p{0.4\linewidth}}
moin@DeutscheKI.de\\
hajo UG, Quarnbek, Germany
\end{tabular}\hspace{0.2in}
$^{\Box}$~\begin{tabular}{p{0.4\linewidth}}
Institute for Neuro- and Bioinformatics,\\
University of Lübeck, Germany
\end{tabular}
}}

\iclrfinalcopy 
\begin{document}

\maketitle

\begin{abstract}
This paper presents TEVR, a speech recognition model
designed to minimize the variation in token entropy w.r.t. to the language model.
This takes advantage of the fact that 
if the language model will reliably and accurately predict a token anyway,
then the acoustic model doesn't need to be accurate in recognizing it.
We train German ASR models with 900~million parameters and show that on CommonVoice German,
TEVR scores a very competitive $3.64\%$ word error rate,
which outperforms the best reported results
by a relative $16.89\%$ reduction in word error rate.
We hope that releasing our fully trained speech recognition pipeline to the community 
will lead to privacy-preserving offline virtual assistants
in the future.

\end{abstract}

\section{Introduction}
\label{section-intro}

Automated speech recognition has improved at an astonishing pace in recent years. This progress has been enabled in part by access to sufficient computing power for training very large models, such as the wav2vec~2.0~XLS-R~1B~\citep{wav2vec2_2020, xlsr_2020, xlsr_1B_2021}, which has 900 million parameters. But the crucial ingredient for success have been more advanced models, which massively reduced the cost for obtaining training data. 

The invention of the CTC Loss \citep{ctc_loss} has made it possible to use unaligned text as the ground-truth signal for speech recognition. Previously, characters and/or phonemes needed to be manually aligned in time to the input audio signal, which was very time consuming. In effect, the CTC Loss has made the preparation of supervised training data $99\%$ cheaper, thereby paving the way for large data-hungry models.

Similarly, the wav2vec 2.0 contrastive loss function has made it possible to use arbitrary speech audio data without any ground-truth text for unsupervised pre-training. The result in practice is that researchers can use the audio from unlabeled videos for pre-training their models up to the point where only a few hours of supervised fine-tuning are necessary to produce excellent results.

Regarding the language models, stochastic techniques such as KenLM's on-disk Kneser-Ney smoothing \citep{kenlm,kenlm-smoothing} have enabled researchers to extract spelling and grammar knowledge out of very large unstructured text collections, such as the OSCAR dataset.

Continuing this tradition, we present TEVR, a stochastic approach for designing recognition tokens based on a large unstructured text collection in such a way that they maximize the information gained by the acoustic speech recognition model without increasing computational complexity or training time.

Our work builds on the pre-trained XLS-R~1B model released by \citet{xlsr_1B_2021}
which itself is a larger variant of XLSR \citep{xlsr_2020}
which is the cross-lingual extension of wav2vec 2.0 \citep{wav2vec2_2020}. 

The model works on partially overlapping 25~ms chunks of audio signal sampled at 16~kHz
which are processed with multiple layers of convolutional embedding 
followed by multiple attention layers to produce a contextual encoding
which is then transformed by a typical linear language head into token logits.
Such an architecture is currently considered state of the art.

Current speech recognition models such as the aforementioned XLS-R~1B are trained using the CTC loss.
The CTC loss minimizes an unweighted sum of the cross-entropies for each time-step. During inference, however, the acoustic model is almost always paired with a stochastic language model. This means that the information gained by correctly recognising a given acoustic token varies strongly based on how this token will later be used by the language model.

To illustrate this concept, consider that "Danke Herr Tajani" and "Dank? Her? Tajani" contain almost the same acoustic information because both unknown tokens indicated by question marks can be reliably and accurately inferred by the language model. Training to accurately recognize these tokens at the acoustic level is, therefore, superfluous. In fact, it might even reduce overall recognition accuracy by taking up resources which otherwise could have learned more useful features. 

In this work, we aim to correct this loss misallocation by introducing multi-character tokens, which are designed to minimize the inter-token variance of the entropies of the combined acoustic and linguistic likelihood distributions used to predict the token sequence representing the recognized text.

\section{Training and Testing Data}

\citet{xlsr_1B_2021} pre-trained the XLS-R~1B model on a total of 436K hours of publicly available audio data from VoxPopuli~\citep{voxpopuli}, Multilingual Librispeech~\citep{mls}, CommonVoice~\citep{commonvoice}, VoxLingua107~\citep{voxlingua107}, and BABEL~\citep{babel}.

We combine the pre-trained convolutional embedding of XLS-R~1B with our own attention encoder and language model and perform additional pre-training matching the instructions in \citet{xlsr_1B_2021}, followed by stochastic language modelling and task-specific fine-tuning.

We perform TEVR token extraction on the training texts from CommonVoice~8.0, Multilingual Librispeech, EuroParl~\citep{eparl}, and OSCAR~\citep{oscar1,oscar2}.

We fine-tune for the German Speech Recognition task using CommonVoice~8.0~\citep{commonvoice}. For easier comparison to literature, e.g. \citet{scribosermo}, we use CommonVoice~6.1~\citep{commonvoice} for testing.

For testing, we generate 4-gram and 5-gram language models based on the texts from CommonVoice~8.0, Multilingual Librispeech, and EuroParl~\citep{eparl}. 

\section{Experimental Setup}

\subsection{TEVR Token Extraction}
\label{section-tevr}

As explained in Section \ref{section-intro}, TEVR tokens are introduced to prevent the loss misallocation caused by uniform weighting of single-character tokens during training of the acoustic model.

\begin{equation}
\label{eq-g}
G_i \sim \delta(t_i-gt_i) = \left\lbrace
\begin{array}{l} 
1 \text{ if } t_i = gt_i \\ 
0 \text{ otherwise} 
\end{array}
\right.
\end{equation}
\begin{equation}
\label{eq-p}
P_i \sim LM(t_0, \mathellipsis, t_{i-1}) 
\end{equation}
\begin{equation}
\label{eq-ce}
\text{lm-entropy}(i) := - \displaystyle\sum_{x \in X} ^{}  \mathbb{P}_{G_i}(x)[\text{log} \mathbb{P}_{P_i}(x)]
\end{equation}
\begin{equation}
\label{eq-ce2}
\text{lm-entropy}(i) = - \text{log} \mathbb{P}_{P_i}(gt_i)
\end{equation}

We define $\text{lm-entropy}(i)$ to mean the sparse categorical cross-entropy for the correct prediction of an unknown token $t_i$ based on the likelihood distribution generated by the ByT5 language model conditioned upon the known tokens $t_0$ to $t_{i-1}$. 

We model the sparse categorical ground-truth distribution for token $ t_i $ at time-step $i$ with \autoref{eq-g}. 
As indicated by the $\sim$ notation, the $G_i$ are considered random variables which are sampled independently from the $\delta(t_i-gt_i)$ distributions.
Correspondingly, the predictions $P_i$ are considered random variables, which are independently sampled for each $i$ from the distributions generated by a fully trained causal deep learning language model $ LM $ conditioned on $t_0$ to $t_{i-1}$, as specified in \autoref{eq-p}.
With set $X$ containing the entire vocabulary from which tokens are selected, we use $\mathbb{P}_{P_i}(x)$ and $\mathbb{P}_{G_i}(x)$ for $x \in X$ to designate the discrete probability density functions of random variables $P_i$ and $G_i$ respectively.
Calculating the cross-entropy of the distribution of $P_i$ relative to the distribution of $G_i$ independently for each time-step $i$ results in \autoref{eq-ce}. Please note that we treat the time index $i$ as a parameter and do not sum the cross-entropies over time.

Due to the sparseness of the distribution of each $G_i$, this can be simplified to \autoref{eq-ce2}, where tokens $gt_i$ are the ground-truth for the language model's prediction for each time-step $i$.
When used without a parameter and in reference to a token, we define lm-entropy to mean $ \text{lm-entropy}(i) $ with $i$ being the index inside the sentence of the current token.
As a result, the lm-entropy of a token is a measure for the amount of information, which the acoustic model needs to generate in order for the language model to accurately predict said token.

\begin{figure}[h]
\begin{center}
\includegraphics[width=\linewidth]{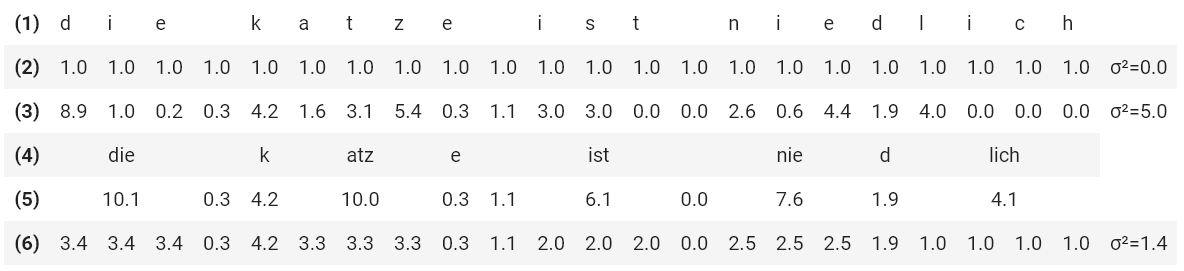}
\end{center}
\caption{Illustration of the entropy variance reduction effect, as explained in Section~\ref{section-tevr}. }
\label{figure-tevr}
\end{figure}

Figure~\ref{figure-tevr} shows each letter of the sentence "Die Katze ist niedlich" in row (1). 
Row (2) shows the entropies of the likelihood distributions generated by the acoustic model for predicting each single-character token when optimized with a stock CTC loss. 
Row (3) shows the per-character lm-entropies, meaning the entropies of the likelihood distributions generated by a stochastic language model for predicting the next single-character token based on knowledge of previous single-character tokens. 
This sequence of lm-entropies observed in the real world has a variance of $\sigma^2=5.0$,
but the CTC loss formula assumes an uniform distribution. 
Obviously, such a discrepancy between theory and practice heavily skews the gradient, thereby optimizing the acoustic model towards the wrong goal.

TEVR tokens greatly alleviate the issue by reducing the variance of the lm-entropies, hence the name "Token Entropy Variance Reduction". This is achieved in three steps:

First and as a prerequisite, we need to calculate the per-character lm-entropies for a large collection of words. Inspired by the suggestion in \citet{byt5} to use raw unicode bytes as tokens for training large T5 \citep{origt5} transformer language models, we determine the lm-entropy of each character by training a T5 model on a large unstructured collection of German texts consisting of roughly 22~billion words. During training, we predict the next character conditioned upon the beginning of the sentence so far and use the cross-entropy as loss signal. We train the model until convergence.

Second, we sample the fully trained model by predicting the next character following each possible prefix of each sentence. We record the sequence of $ \text{lm-entropy}(c=\text{Groundtruth}_i,i)\ \forall i $ and obtain the values shown in Figure~\ref{figure-tevr} row (3). For example, the partial word "niedl???" can be reliably corrected to "niedlich" by the language model, since the "i~c~h" characters have low lm-entropies in row (3).

Third, we then use these per-character lm-entropies to extract compound tokens. For each character and each sentence in the CommonVoice training split we iterate over all sub-strings to find the lowest-lm-entropy snippets of a given length. For each sentence, we only retain the $20\%$ lowest-lm-entropy snippets. We then select the most-common snippets over all sentences as our TEVR tokens for any given length. 

Our model variant M consists of 40 4-character tokens, 80 3-character tokens, 96 2-character tokens, and the 40 single-character tokens used in variant S.
For lists of the specific tokens used for each model variant, please see the appendix or our source code release at\\
\url{https://huggingface.co/fxtentacle/wav2vec2-xls-r-1b-tevr}.

Row (4) of Figure~\ref{figure-tevr} shows the tokenization of the example sentence using the TEVR tokens from model variant M. 
Row (5) shows the per-token lm-entropies for these compound character tokens, as obtained by summing up the individual per-character lm-entropies of contained characters. 
Row (6) then shows the effective per-character lm-entropies used for gradient estimation when the model variant M is trained with TEVR tokens. With $\sigma^2=1.4$, the lm-entropy sequence of these tokens is much closer to the uniform distribution assumed by the CTC loss, thereby greatly reducing the gradient skew during training.

To highlight that the performance improvement of TEVR is caused by the lm-entropy variance reduction and not just by introducing multi-character tokens, we also train and evaluate model variant L which contains almost all multi-character tokens seen in any training sentence, but chosen exhaustively and, hence, without considering the per-character lm-entropy.

\subsection{German Speech Recognition}

We use the wav2vec 2.0 implementation available from \citet{huggingface}
with a configuration of 48 hidden layers using 1280 hidden states each
and feed-forward blocks with an inner dimension of 5120.
Together with the LM head, that results in 918~million to 919~million parameters 
depending on variant.
We use gradient checkpointing \citep{checkpointing} to conserve GPU memory,
and the Adam optimizer \citep{adam} with a learning rate warm-up followed by linear decay. Due to memory constraints, we exclude audio recordings that exceed a length of 295,488 samples. Each training batch then consists of 8 complete audio recordings. A 0.005~weigh~decay was used.

We train 3 variants of the model for up to 4 epochs each with varying warm-up and learning rates, as summarized in Table~\ref{table-train} and Table~\ref{table-variants}.
\\

\begin{table}[h]
\begin{center}
\begin{tabular}{lll}
Epoch & Warm-up Steps & Learning Rate \\
\hline \\
1 & 500 & $1.0 \cdot 10^{-4}$ \\
2 & 1500 & $0.5 \cdot 10^{-4}$ \\
3 & 1500 & $0.1 \cdot 10^{-4}$ \\
4 & 1500 & $0.1 \cdot 10^{-4}$ \\
\end{tabular}
\end{center}
\caption{Training Schedule}
\label{table-train}
\end{table}

\begin{table}[h]
\begin{center}
\begin{tabular}{ l p{0.1\linewidth} p{0.12\linewidth} p{0.58\linewidth} }
Variant & Number of Tokens & Total \nobreak{Parameters} & Notes \\
\hline \\
S & 40 & 918~million & Unmodified wav2vec 2.0 XLS-R 1B architecture with single character tokens, as used by \citet{xlsr_1B_2021} in their evaluation. \\
M & 256 & 918~million & In addition to all single-character tokens from variant S, 40 TEVR 4-character tokens, 80 TEVR 3-character tokens, and 96 TEVR 2-character tokens were added.  \\
L & 904 & 919~million & This variant contains the majority of all 4-character, 3-character, and 2-character sub-strings, thereby nullifying the variance reduction effect. 
\end{tabular}
\end{center}
\caption{Model Variants. See Section~\ref{section-tevr} for an explanation of how the tokens were obtained.}
\label{table-variants}
\end{table}

\begin{table}[p]
\begin{center}
\begin{tabular}{lccc}
Epoch & WER Variant S & WER Variant M & WER Variant L \\
\hline \\
1 & $14.83\%$ & $\bm{14.59\%}$ & $15.10\%$ \\
2 & $12.69\%$ & $\bm{12.25\%}$ & $12.72\%$ \\
3 & $11.38\%$ & $\bm{10.81\%}$ & $11.21\%$ \\
4 & $10.72\%$ & $\bm{10.10\%}$ & $10.53\%$ 
\end{tabular}
\end{center}
\caption{Raw Word Error Rates for each model variant after each epoch. These error rates were calculated without a language model. The best result in each epoch is highlighted in bold.}
\label{table-results-raw-wer}
\end{table}

\begin{table}[p]
\begin{center}
\begin{tabular}{lccc}
Epoch & CER Variant S & CER Variant M & CER Variant L \\
\hline \\
1 & $\bm{3.74\%}$ & $4.10\%$ & $4.54\%$ \\
2 & $\bm{3.18\%}$ & $3.42\%$ & $3.76\%$ \\
3 & $\bm{2.82\%}$ & $2.96\%$ & $3.27\%$ \\
4 & $\bm{2.64\%}$ & $2.78\%$ & $3.06\%$ 
\end{tabular}
\end{center}
\caption{Raw Character Error Rates for each model variant after each epoch. These error rates were calculated without a language model. The best result in each epoch is highlighted in bold.}
\label{table-results-raw-cer}
\end{table}

\begin{table}[p]
\begin{center}
\begin{tabular}{cc ccc}
Alpha & Beta & WER Variant S & WER Variant M & WER Variant L \\
\hline \\
0.5 & 0.5 & $5.95\%$ & ${4.93\%}$ & $5.95\%$ \\
0.6 & 0.5 & $5.79\%$ & ${4.93\%}$ & $5.84\%$ \\
0.7 & 0.5 & $5.59\%$ & ${4.73\%}$ & $5.54\%$ \\
0.8 & 0.5 & $5.95\%$ & ${4.88\%}$ & $5.54\%$ \\
0.5 & 0.75 & $5.95\%$ & ${4.98\%}$ & $5.84\%$ \\
0.6 & 0.75 & $5.69\%$ & ${4.93\%}$ & $5.79\%$ \\
\textbf{0.7} & \textbf{0.75} & $5.59\%$ & $\bm{4.73\%}$ & $5.59\%$ \\
0.8 & 0.75 & $5.69\%$ & ${4.83\%}$ & $5.54\%$ \\
0.5 & 1.0 & $5.79\%$ & ${4.93\%}$ & $5.74\%$ \\
0.6 & 1.0 & $5.39\%$ & ${4.93\%}$ & $5.79\%$ \\
0.7 & 1.0 & $5.34\%$ & ${4.73\%}$ & $5.64\%$ \\
0.8 & 1.0 & $5.64\%$ & ${4.83\%}$ & $5.54\%$ 
\end{tabular}
\end{center}
\caption{Parameter sweep on a validation split for identifying the best $\alpha$ and $\beta$ parameters for decoding with a 4-gram language model. For every set of parameters, model variant M performed significantly better than variant S and L. Our final parameter selection of $\alpha=0.7, \beta=0.75$  is highlighted in bold.}
\label{table-results-4g-wer}
\end{table}

\begin{table}[p]
\begin{center}
\begin{tabular}{l l l l}
Architecture & Language Model & WER & Source \\
\hline \\
wav2vec 2.0 XLS-R 1B + TEVR & 5-gram & $\bm{3.64\%}$ & our best result \\
wav2vec 2.0 XLS-R 1B + TEVR & 4-gram & $3.70\%$ & our ablation 4-gram LM \\
wav2vec 2.0 XLS-R 1B & 5-gram & $4.38\%$ & \citet{flozi} * \\
QuartzNet15x5DE (D37) & 5-gram & $6.6\%$ & \citet{scribosermo} \\
wav2vec 2.0 XLS-R 1B + TEVR & no LM & $10.10\%$ & our ablation no LM \\
wav2vec 2.0 XLS-R & no LM & $12.06\%$ & \citet{grosman} * \\
\end{tabular}
\end{center}
\caption{Overview of our word error rates and results from literature on German CommonVoice~6.1. The entries marked with * indicate self-reported results on the Community Models tab on paperswithcode.com. We successfully replicated their results based on the source code provided by them.}
\label{table-results-final-wer}
\end{table}

\FloatBarrier

\section{Results}

The raw word error rates without language model obtained by each of the 3 variants after each of the 4 training epochs are shown in Table~\ref{table-results-raw-wer}. After 4 epochs, variant M with TEVR tokens reduces the raw word error rate by a relative $5.74\%$ in comparison to variant S, which is the unmodified wav2vec 2.0 XLS-R 1B architecture. 
As will be shown later, this performance improvement is amplified when using a word model.
For character error rates, see Table~\ref{table-results-raw-cer}. Unsurprisingly, using single character tokens resulted in the lowest raw character error rate while the model variant with the highest number of distinct tokens also had the highest error rate.

We performed a parameter sweep using a small 4-gram word model on a validation split of the data, which confirmed our hypothesis that the TEVR-enhanced model variant M performs best, this time with a relative $15.38\%$ improvement in word error rate over the unmodified wav2vec 2.0 XLS-R 1B architecture. Please also note that model variant L, which contains even more multi-character tokens, performs worse than variant M. This strongly suggest that the performance improvement is indeed caused by the entropy variance reduction and not merely by introducing multi-character tokens. The results are provided in Table~\ref{table-results-4g-wer}. 

We chose model variant M with 4 epochs of training as the acoustic model and built a full recognition pipeline (including language model) based on it for our final evaluation on  the German testing split of CommonVoice~6.1. We obtained a very competitive word error rate of $3.64\%$ with a 5-gram language model ($\alpha = 0.7$, $\beta = 0.75$).

For comparison, we consulted the top results from paperswithcode.com (retrieved 02.06.2022), which were \citet{scribosermo} ($6.6\%$ WER) in the literature and an unmodified wav2vec 2.0 XLS-R 1B architecture self-reported as a community model by \citet{flozi}.
Accordingly, instead of attempting to build our own unmodified wav2vec 2.0 XLS-R 1B recognition pipeline for comparison, we successfully verified the community results of \citet{flozi} ($4.38\%$ WER with LM) and \citet{grosman} ($12.06\%$ WER without LM). 
See Table~\ref{table-results-final-wer} for a comparison.

\section{Conclusion}

We have shown that when combined with an appropriately tuned language model, the TEVR-enhanced model outperforms the best German automated speech recognition result from literature by a relative $44.85\%$ reduction in word error rate. 
It also outperforms the best self-reported community model by a relative $16.89\%$ reduction in word error rate.

We have shown that TEVR tokens outperform generically chosen multi-character tokens of a similar length, which suggests that it is the entropy variance reduction technique that leads to the increased model performance. 
We observe that the TEVR tokens - which were chosen with the goal of reducing entropy variance -  coincide with linguistically meaningful word endings, such as "kätz-chen", "nied-lich", "funk-tion", and "glaub-haft". This hints at redundancy in the German language which TEVR tokens can exploit, but single-character tokens such as those used in the unmodified wav2vec~2.0~XLS-R~1B~architecture cannot. See the appendix for a full list of the tokens used for all 3 model variants.

Due to budget constraints, we did not attempt to train a full TEVR-enhanced speech recognition pipeline for English. For professional clean studio-quality audiobook recordings, such as those in the LibriSpeech English dataset, we would expect TEVR to yield only a low relative improvement, because in those situations there are only few acoustic ambiguities. When operating under difficult conditions, such as the real-world dictation examples in the CommonVoice English dataset, however, we would expect TEVR to also significantly improve English ASR performance. We plan to explore this in a future work.

Our final German speech recognition pipeline consisting of the acoustic model, a matching language model, and necessary source code for inference and evaluation will be made publicly available on:

\url{https://huggingface.co/fxtentacle/wav2vec2-xls-r-1b-tevr}

We're hoping to see privacy-preserving offline virtual assistants in the future and we hope that releasing a fully trained working German ASR pipeline will help to make progress towards that goal.

\bibliography{paper}
\bibliographystyle{iclr2022_conference}

\newpage
\section{Appendix}

\subsection{Tokens used by Model Variant S}
This is an unmodified wav2vec~2.0~XLS-R~1B~architecture with single-character tokens:

e, n, t, h, r, i, s, d, g, l, c, a, u, m, k, f, o, z, b, w, p, v, ä, ü, j, ö, y, q, x, í, č, ō, æ, á, ś, š, ?

\subsection{Tokens used by Model Variant M}
These are the TEVR tokens obtained by grouping multiple low-lm-entropy characters into medium-lm-entropy compound tokens:

chen, sche, lich, isch, icht, iche, eine, rden, tion, urde, haft, eich, rung, chte, ssen, chaf, nder, tlic, tung, eite, iert, sich, ngen, erde, scha, nden, unge, lung, mmen, eren, ende, inde, erun, sten, iese, igen, erte, iner, tsch, keit, der, die, ter, und, ein, ist, den, ten, ber, ver, sch, ung, ste, ent, ach, nte, auf, ben, eit, des, ers, aus, das, von, ren, gen, nen, lle, hre, mit, iel, uch, lte, ann, lie, men, dem, and, ind, als, sta, elt, ges, tte, ern, wir, ell, war, ere, rch, abe, len, ige, ied, ger, nnt, wei, ele, och, sse, end, all, ahr, bei, sie, ede, ion, ieg, ege, auc, che, rie, eis, vor, her, ang, für, ass, uss, tel, er, in, ge, en, st, ie, an, te, be, re, zu, ar, es, ra, al, or, ch, et, ei, un, le, rt, se, is, ha, we, at, me, ne, ur, he, au, ro, ti, li, ri, eh, im, ma, tr, ig, el, um, la, am, de, so, ol, tz, il, on, it, sc, sp, ko, na, pr, ni, si, fe, wi, ns, ke, ut, da, gr, eu, mi, hr, ze, hi, ta, ss, ng, sa, us, ba, ck, em, kt, ka, ve, fr, bi, wa, ah, gt, di, ab, fo, to, rk, as, ag, gi, hn, s, t, n, m, r, l, f, e, a, b, d, h, k, g, o, i, u, w, p, z, ä, ü, v, ö, j, c, y, x, q, á, í, ō, ó, š, é, č, ?

\subsection{Tokens used by Model Variant L}
This is an almost exhaustive list of all multi-character sequences seen in the training data and was chosen without regards to lowering lm-entropy variance:

aben, ache, acht, afte, aftl, agen, ahme, ahre, alen, alle, alls, alte, altu, amen, amme, ande, anis, annt, asse, aten, atet, atio, ativ, atte, auch, beit, biet, chaf, chen, cher, ches, chie, chke, chla, chli, chne, chst, chte, chti, chts, chtu, chun, ckel, cklu, delt, dene, dent, dere, dern, ders, dert, deru, dete, dier, dies, dige, ding, dlic, dlun, dnet, doch, dung, eben, eche, echt, eden, eder, edoc, egen, ehen, ehme, ehör, eich, eide, eind, eine, eise, eist, eite, eits, elen, eler, elle, ellt, elte, ende, enen, ensc, ente, entl, erde, erem, eren, erli, erne, ersc, ersi, erst, erte, erun, esch, esem, esen, eser, esse, essl, este, eten, eter, etzt, eute, fall, fent, ffen, ften, ftig, ftli, gend, gene, gisc, gkei, glic, glie, grei, gten, gung, haft, halb, heit, hend, hied, hkei, hlan, hlic, hlie, hlre, hmen, hnet, hnun, hren, hrer, hrte, hrun, hsel, hste, hten, hter, htet, htig, htun, hule, hung, hört, iche, ichk, ichn, icht, iden, iebe, iede, iegt, iele, iell, ielt, iere, iert, iese, iess, igen, iger, igke, igte, igun, ilie, inde, inem, inen, iner, ines, inge, ings, inie, insa, iona, ione, irat, isch, isse, issi, iste, iten, iter, itet, itig, itik, itio, itis, itte, ität, jahr, kann, keit, klun, krie, ktio, kung, lame, land, ldet, ldun, lich, lied, lien, lies, lige, lisc, liti, llem, llen, llte, llun, logi, lrei, lsch, lten, ltkr, ltun, lung, luss, mati, mein, ment, miss, mmen, mung, nahm, nale, nand, nden, nder, ndes, ndet, ndig, ndun, nete, nfal, ngen, nger, ngli, nich, nier, nige, nisc, nlic, nnen, nnte, nsam, nsch, nten, nter, ntli, nung, olge, olgt, omme, onal, onde, onen, opäi, oren, osse, piel, päis, rach, rate, rati, rbei, rche, rdem, rden, rdin, reic, rend, rere, rger, rhei, rich, rieb, rieg, rier, rige, risc, ritt, rlic, ropa, ropä, rsch, rsit, rste, rten, rund, rung, scha, sche, schl, send, serd, sere, sich, side, sier, sind, sion, sisc, sitä, slic, sond, spie, ssen, sser, ssig, ssio, ssli, sste, sten, ster, sung, tadt, tand, teht, teil, tell, tere, tern, ters, tete, tier, tige, tigt, tion, tisc, tive, tkri, tlic, trie, tsch, ttel, tten, tter, tung, tzen, tzte, tzun, unge, ungs, unkt, urch, urde, utsc, utun, vers, weis, weit, werd, wird, wort, wurd, zeit, ziel, zier, zlic, zten, zung, ächs, äisc, ändi, äter, öcht, önne, örte, über, ührt, üngl, üsse, abe, ach, age, ahl, ahn, ahr, akt, ale, ali, all, als, alt, ame, amt, and, ang, ank, ann, ans, ant, anz, ara, arb, are, ari, ark, art, ass, ast, ate, atz, aub, aue, auf, aus, aut, bar, bau, bed, bef, beg, bei, bek, bel, ben, ber, bes, bet, bew, bez, bil, bis, ble, bli, bra, bst, bur, cha, che, chi, chr, chs, chw, cke, dan, dar, das, dem, den, der, des, deu, die, dor, dre, ebt, eck, ege, egi, ehr, eil, eim, ein, eis, eit, ekt, ele, ell, elt, eme, enn, ens, ent, era, erb, ere, erf, erg, erh, eri, erk, erl, ern, err, ers, ert, erw, erz, est, etr, eue, eut, fel, fen, fer, ffe, for, fra, fre, für, gan, geb, gef, geh, gel, gem, gen, ger, ges, gew, gib, gra, han, hat, hau, hei, her, hie, hin, hle, hne, hre, ial, ibt, ich, ick, iel, ien, ier, iff, ige, ihn, ihr, ill, imm, ina, ind, ine, ing, ini, inn, ins, inz, ion, ist, jah, ken, ker, kom, kon, kte, lag, lan, lau, leg, lei, len, ler, lie, lig, lis, lle, mal, man, mar, meh, mei, men, mer, mis, mit, mmt, mte, mus, nac, nde, nen, ner, net, neu, nge, nie, nis, nke, nne, noc, nor, nst, ntr, nur, nze, obe, och, ode, ohn, oll, ord, org, ori, orm, ort, par, pen, per, pla, pol, ppe, pre, pro, rag, ran, rat, rau, reg, rei, ren, res, rge, rie, rif, rin, ris, rke, rla, rte, rts, rze, rüc, sam, sch, seh, sei, sel, sen, ser, sge, sie, sol, son, spr, sst, sta, ste, sti, str, stu, stä, tal, tar, tei, tel, ten, ter, the, tig, tor, tra, tre, tri, tro, tru, tsc, tur, ube, uch, ude, uen, uer, und, ung, uni, uns, upt, urg, usg, uss, ust, ute, ver, vie, vol, vom, von, vor, war, was, wei, wel, wen, wer, wie, wir, woh, zei, zen, zum, zur, zus, zwe, ück, ab, ad, ag, ak, al, am, an, ar, as, at, au, ba, be, bi, bo, br, bu, ch, da, de, di, do, du, eb, ed, eg, eh, ei, el, em, en, er, es, et, eu, fa, fe, fi, fl, fr, ft, fü, ga, ge, gi, gl, gr, gt, ha, he, hi, ho, hä, ia, id, ie, ig, ih, ik, il, im, in, ir, is, it, iv, je, ka, ke, ki, kl, ko, kr, ku, kö, la, ld, le, li, lo, lt, lu, lä, ma, me, mi, mo, mp, mu, mü, na, nd, ne, nf, ng, ni, no, nt, of, ol, om, on, op, or, os, ot, pa, pf, pl, po, pr, ra, rd, re, ri, ro, rt, ru, rä, rü, sa, sc, se, si, so, sp, ss, st, ta, te, th, ti, to, tr, ts, tt, tu, tz, uf, ug, ul, um, un, ur, us, ut, ve, vo, wa, we, wi, wo, wu, ze, zi, zu, äh, äu, ün, ür, a, b, c, d, e, f, g, h, i, j, k, l, m, n, o, p, q, r, s, t, u, v, w, x, y, z, á, ä, é, í, ó, ö, ü, ł, ō, š, ?

\end{document}